\documentclass[10pt,twocolumn,letterpaper]{article}

\usepackage[pagenumbers]{cvpr} 

\usepackage{graphicx}
\usepackage{amsmath}
\usepackage{amssymb}
\usepackage{booktabs}
\usepackage[table,xcdraw]{xcolor}
\usepackage{stfloats}
\usepackage{physics}
\usepackage[accsupp]{axessibility}  

\usepackage[pagebackref,breaklinks,colorlinks]{hyperref}

\usepackage[capitalize]{cleveref}
\crefname{section}{Sec.}{Secs.}
\Crefname{section}{Section}{Sections}
\Crefname{table}{Table}{Tables}
\crefname{table}{Tab.}{Tabs.}

\def\cvprPaperID{8067} 
\def\confName{CVPR}
\def\confYear{2023}

\begin{document}

\title{HOICLIP: Efficient Knowledge Transfer for HOI Detection with Vision-Language Models}


\author{Shan Ning\textsuperscript{\rm1}\footnotemark[1] \space\space
Longtian Qiu\textsuperscript{\rm1}\footnotemark[1] \space\space
Yongfei Liu\textsuperscript{\rm2} \space\space
Xuming He\textsuperscript{\rm1,3}\\
ShanghaiTech University, Shanghai, China\textsuperscript{\rm1} \space\space\space ByteDance Inc.\textsuperscript{\rm2}  \\
Shanghai Engineering Research Center of Intelligent Vision and Imaging\textsuperscript{\rm3} \\ 
{\tt\small {\{ningshan2022, qiult, hexm\}@shanghaitech.edu.cn \space\space liuyongfei314@gmail.com}}
\and 
}
\maketitle

\renewcommand{\thefootnote}{\fnsymbol{footnote}}
\footnotetext[1]{These authors contributed equally to this work, which was supported by Shanghai Science and Technology Program 21010502700 and Shanghai Frontiers Science Center of Human-centered Artificial Intelligence.}



\begin{abstract}
   Human-Object Interaction (HOI) detection aims to localize human-object pairs and recognize their interactions. Recently, Contrastive Language-Image Pre-training (CLIP) has shown great potential in providing interaction prior for HOI detectors via knowledge distillation. However, such approaches often rely on large-scale training data and suffer from inferior performance under few/zero-shot scenarios. In this paper, we propose a novel {HOI} detection framework that efficiently extracts prior knowledge from CLIP and achieves better generalization. In detail, we first introduce a novel interaction decoder to extract informative regions in the visual feature map of CLIP via a cross-attention mechanism, which is then fused with the detection backbone by a knowledge integration block for more accurate human-object pair detection. In addition, prior knowledge in CLIP text encoder is leveraged to generate a classifier by embedding HOI descriptions. To distinguish fine-grained interactions, we build a verb classifier from training data via visual semantic arithmetic and a lightweight verb representation adapter. Furthermore, we propose a training-free enhancement to exploit global HOI predictions from CLIP. Extensive experiments demonstrate that our method outperforms the state of the art by a large margin on various settings, e.g. +4.04 mAP on HICO-Det. The source code is available in \href{https://github.com/Artanic30/HOICLIP}{https://github.com/Artanic30/HOICLIP}.
   
\end{abstract}

\section{Introduction}
\label{sec:intro}


Human-Object Interaction (HOI) detection, which aims to localize human-object pairs and identify their interactions, is a core task towards a comprehensive understanding of visual scenes. It has attracted increasing interest in recent years for its key role in a wide range of applications, such as assistive robots, visual surveillance and video analysis \cite{bemelmans2012socially,dee2008close,feichtenhofer2017spatiotemporal,bolme2010visual}. 
Thanks to the development of end-to-end object detectors\cite{detr}, recent research\cite{qpic, gen, CDN,hotr,XinpengLiu2022InteractivenessFI} has made remarkable progress in localizing human-object instances in interaction. 
Nonetheless, the problem of identifying interaction classes between human-object pairs remains particularly challenging. Conventional strategies \cite{as-net,hotr,qpic,CDN} simply learn a multi-label classifier and typically require large-scale annotated data for training. As such, they often suffer from long-tailed class distributions and a lack of generalization ability to unseen interactions.


\begin{figure}
     \centering
     \begin{subfigure}[t]{0.23\textwidth}
         \centering
         \includegraphics[width=\textwidth]{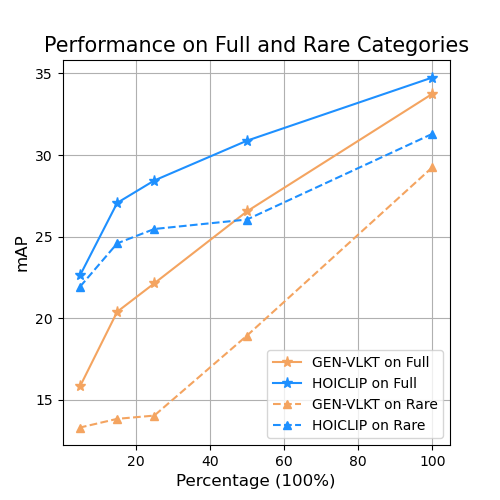}
         \caption{Data Efficiency Comparison}
     \end{subfigure}
     \hfill
     \begin{subfigure}[t]{0.23\textwidth}
         \centering
         \includegraphics[width=\textwidth]{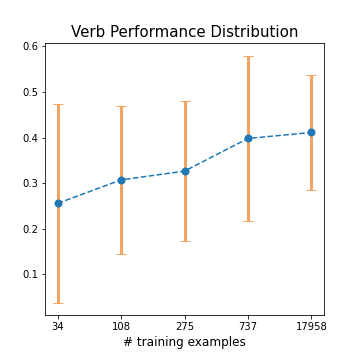}
         \caption{Verb Long-tail Problem}
         \label{fig:frac_rare}
     \end{subfigure}
     \vspace{-0.3cm}
        \caption{\textbf{Data efficiency comparison and verb distribution analysis.} In panel (a), we increase training data from $5\%$ to $100\%$ and show the result of HOICLIP and GEN-VLKT. In panel (b), the dots indicate the mean mAP and length of vertical line indicate the variance of mAP for verbs grouped by sample number.}
\label{fig:frac}
\vspace{-0.4cm}
\end{figure}

Recently, Contrastive Vision-Language Pre-training\cite{clip} has been explored to address such open-vocabulary and zero-shot learning problems as its learned visual and linguistic representations demonstrate strong transfer ability in various downstream tasks.  
In particular, recent work on open-vocabulary detection utilizes knowledge distillation to transfer CLIP's object representation to object detectors\cite{vild, prompt_vild,zhao2022exploiting,feng2022promptdet,ma2022open,DBLP:journals/corr/abs-2109-12066}. Such a strategy has been adopted in the work of HOI detection, including  GEN-VLKT\cite{gen} and EoID\cite{eoid}, which leverage CLIP's knowledge to tackle the long-tail and zero-shot learning in the HOI tasks.

Despite their promising results, it remains an open question how we effectively transfer CLIP knowledge to the HOI recognition task it involves compositional concepts composed of visual objects and interactions. First, as pointed out in\cite{cho2019efficacy,SamuelStanton2021DoesKD}, the commonly-adopted teacher-student distillation objective is not aligned with improving the generalization of student models. In addition, as shown in Figure~\ref{fig:frac}, we empirically observe that the knowledge distillation in learning HOIs (e.g., GEN-VLKT) typically requires a substantial amount of training data, which indicates its \textit{low data efficiency}. Furthermore, knowledge distillation often suffers from \textit{performance degradation in zero-shot generalization} as it lacks training signal for unseen classes which is critical to inherit knowledge from the teacher model.



To address those challenges, we propose a novel strategy, dubbed HOICLIP, for transferring CLIP knowledge to the HOI detection task in this work. Our design ethos is to directly retrieve learned knowledge from CLIP instead of relying on distillation and to mine the prior knowledge from multiple aspects by exploiting the compositional nature of the HOI recognition. Moreover, to cope with the long-tail and zero-shot learning in verb recognition under low data regime, we develop a verb class representation based on visual semantic arithmetic, which does not require large amount of training data as in knowledge distillation based methods. Our methods enable us to improve the data efficiency in HOI representation learning and achieve better generalization as well as robustness.


Specifically, our HOICLIP framework learns to retrieve the prior knowledge from the CLIP model from three aspects: 
1) \textit{Spatial feature}. 
As the feature location is key to the detection task, we fully exploit the visual representation in CLIP and extract features only from informative image regions. 
To this end, we utilize CLIP's feature map with spatial dimensions and develop a transformer-based interaction decoder that learns a localized interaction feature with cross-modal attention.     
2) \textit{Verb feature}.
To address the long-tailed verb-class problem as shown in Figure \ref{fig:frac}, we develop a verb classifier focusing on learning a better representation for the verbs. Our verb classifier consists of a verb feature adapter\cite{adapter,vladapter,clipadapter,tipadapter} and a set of class weights computed via visual semantic arithmetic \cite{zero_cap}. We enhance the HOI prediction by fusing the outputs of the verb classifier and the common interaction classifier.  
3) \textit{Linguistic feature}. To cope with the very rare and unseen class for HOI prediction, we adopt a prompt-based linguistic representation for HOIs and build a zero-shot classifier for the HOI classification\cite{MitchellWortsman2021RobustFO}. This classifier branch requires no training and we integrate its output with the HOI classifier during model inference. 


We evaluate our HOICLIP on two representative HOI detection datasets, HICO-DET\cite{hico_det} and V-COCO\cite{vcoco}. To validate HOICLIP, we perform extensive experiments under fully-supervised setting, zero-shot setting and data-efficient setting. The experiment results demonstrate the superiority of our methods: HOICLIP achieves competitive performance across all three settings, outperforming previous state-of-the-art methods on the zero-shot setting by 4.04 mAP and improving the data efficiency significantly.

The main contributions of our paper can be summarized as follows:
\begin{itemize}
\setlength{\itemsep}{0pt}%
\setlength{\parskip}{0pt}%
    \item To our best knowledge, HOICLIP is the \textbf{first work} to utilize query-based knowledge retrieval for efficient knowledge transfer from the pre-trained CLIP model to HOI detection tasks.
    \item We develop a fine-grained transfer strategy, leveraging regional visual features of HOIs via cross-attention and a verb representation via visual semantic arithmetic for more expressive HOI representation. 
    \item We further improve the performance of HOICLIP by exploiting zero-shot CLIP knowledge without additional training.
\end{itemize}

\section{Related work}
%
\label{sec:related}
\textbf{HOI Detection.} The HOI detection task mainly involves three sub-problems, including object detection, human-object pairing and interaction recognition. Previous HOI detection methods can be categorized into two-stage and one-stage paradigm. The two-stage \cite{IDN,VSG_net,zhong2020polysemy,li2020detailed,kim2020detecting,wan2023weakly} paradigm methods use an independent detector to obtain locations and classes of objects, followed by specifically-designed modules for human-object association and interaction recognition. A typical strategy is to use graph-based methods to extract relation information to support interaction understanding\cite{VSG_net, DongmingYang2020AGI}. 
The one-stage paradigm instead detects the human-object pairs with interaction directly without a need for stage-wise processing. 
Recently, several HOI methods inspired \cite{hotr,qpic,CDN,XinpengLiu2022InteractivenessFI,gen,eoid,zhong2022towards,wu2022mining,yuanrlip,PenghaoZhou2019RelationPN} by Transformer-based Detectors\cite{detr} have achieved promising performance. In particular,  
GEN-VLKT\cite{gen} further designs a two-branch pipeline to provide a parallel forward process, and uses separated query for human and object instead of the unified query used in CDN. RLIP\cite{yuan2022rlip} propose a pre-training strategy for HOI detection based on image captions. Our method builds on the top of the transformer-based HOI detection strategy and focuses on improving interaction recognition.


\begin{figure*}[t!]
  \centering
    \includegraphics[width=0.925\textwidth]{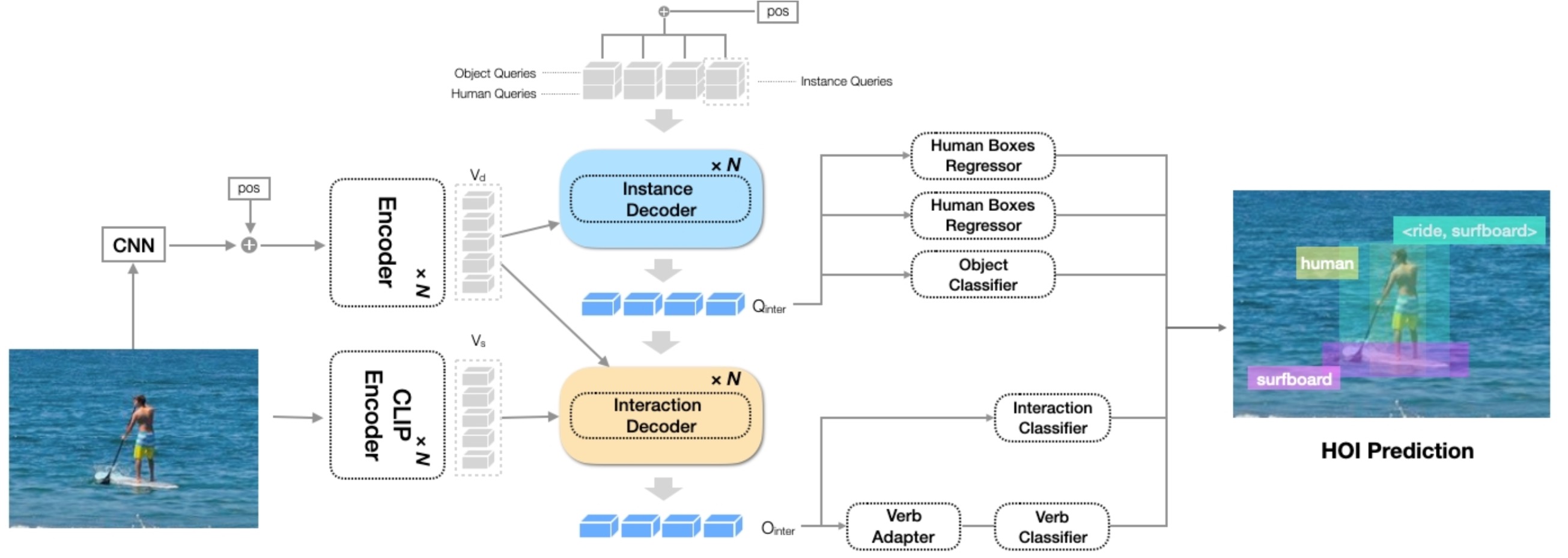}
   \caption{\textbf{Architecture of HOICLIP.} Given an image, HOICLIP encodes it with a detection encoder and CLIP encoder. The instance decoder localizes human and object pairs using features from the detection encoder. The interaction decoder leverages features from both the encoder and extract interaction representation. The verb adapter extracts verb representation based on the interaction representation.}
    \label{figure:pipeline}
    \vspace{-0.2cm}
\end{figure*}

\textbf{Exploiting Vision-language Models.}
Recent breakthroughs in Vision-Language Models (VLM)\cite{clip, align} demonstrate a promising transfer ability to downstream tasks. The visual representations learned from natural language supervision pave the way for zero-shot and open vocabulary tasks\cite{OVDDETR, vild, prompt_vild, must, zero_cap,zhao2022exploiting,feng2022promptdet,ma2022open,DBLP:journals/corr/abs-2109-12066}. Pioneer works\cite{vild} transfer the VLMs to open vocabulary object detection
 through knowledge distillation. Inspired by this idea, recent research\cite{gen, eoid} adopts the same strategy for HOI detection. Previous efforts to transfer VLM to detection tasks can be summarized into two aspects: (1) Prior knowledge integration through texts, which initializes classifiers with labels' text embedding from CLIP; (2) Feature (or logtis) level knowledge distillation, which guides the learned features (or logit predictions) to align with image feature embedded by CLIP (or logtis predicted by zero-shot CLIP). In this work, we propose a novel strategy for transferring VLM knowledge to HOI detection tasks. Different from the above methods, we directly retrieve related information from CLIP, leading to superior performance and higher data efficiency.
 
\textbf{Zero-shot HOI Detection.} The target of zero-shot HOI detection is to detect and recognize HOI categories absent from training data. Due to the compositionality of HOIs, annotations of all possible HOI combinations are impractical. Therefore, the zero-shot HOI detection setting is important for application in real-world scenarios. Previous work\cite{AnkanBansal2019DetectingHI,gupta2019no,hou2021affordance,VCL,FCL, ConsNet,peyre2019detecting} tackle such a challenge in a compositional manner, which disentangle reasoning on actions and objects during training. This makes it possible to recognize unseen $\langle\text{human, object, verb}\rangle$ combinations during inference. Due to breakthroughs in VLMs\cite{clip}, recent research\cite{gen, eoid} focuses on transferring knowledge from VLMs to recognize unseen HOI concepts and achieve a promising performance gain on the zero-shot setting. Our work aims to explore a more efficient multi-facet strategy for knowledge transfer from VLMs in the zero-shot HOI.
 
\section{Method}
In this section, we introduce our HOICLIP framework for efficient CLIP knowledge transfer to HOI detection and preserving the generalization ability. We depict the overall architecture of our model in Section \ref{sec:arch}, followed by three key aspects of our transfer method. In Section \ref{sec:query}, we introduce the query-based knowledge retrieval strategy for efficient visual knowledge transfer. In Section \ref{sec:verb}, we present our verb representation adapter and verb classifier extraction for verb knowledge transfer. In Section \ref{sec:free}, we develop a training-free enhancement for visual-linguistic knowledge transfer. Finally in Section \ref{sec:train}, we describe our training and inference pipeline., which


\subsection{Overall Architecture}
\label{sec:arch}
The overall architecture of our HOICLIP is illustrated in Figure~\ref{figure:pipeline}. We first adopt the transformer-based end-to-end object detector\cite{detr} to localize the humans and objects. Specifically, given an input image $I$, we use a transformer encoder to obtain a spatial image feature map $V_{d}$, 
followed by instance decoder and interaction decoder to accomplish instance detection and interaction recognition, respectively. Inspired by GEN-VLKT\cite{gen}, the instance decoder takes two groups of queries as the input for human and object respectively, namely human query $Q_h$,
and object query $Q_o$. 
The output object queries $O_o \in R^{N_q \times C_{e}}$ and human queries $O_h \in R^{N_q \times C_{e}}$ in the last decoder layer are used to predict human bounding box $B_h \in R^{N_q \times 4}$, object bounding box $B_o \in R^{N_q \times 4}$ and object class $C_o \in R^{N_q \times K_o}$, where $K_o$ is the number of object classes. 

Given the human and object features, we then introduce a novel interaction decoder to perform interaction recognition, in which we utilize the information from the previous extracted feature map $V_d$ and from a spatial feature map $V_s$ generated by CLIP, and perform a knowledge integration via a cross-attention module. Subsequently, a verb adapter extracts the action information to augment the interaction representation and recognition. A linear classifier takes the output of the interaction decoder to predict the HOI category, which is further enhanced by a training-free classifier using CLIP's linguistic features.

\setlength{\abovedisplayskip}{6pt}
\setlength{\belowdisplayskip}{6pt}

\subsection{Query Based Knowledge Retrieval}
\label{sec:query}
In this part, we describe the design of query-based interaction knowledge retrieval starting from revisiting the pipeline of Zero-Shot CLIP image classification. 
\vspace{-2mm}
\paragraph{Zero-shot CLIP} 
CLIP extract dual-modality features by a visual encoder and text encoder. The visual encoder consists of a backbone $\mathrm{VisEnc}(\cdot)$ and a projection layer $\mathrm{Proj}(\cdot)$. The visual backbone extracts visual spatial feature $V_{s} \in R^{H_s \times W_s \times C_{s}}$, which is fed into the projection layer to obtain a global visual feature $V_{g}\in R^{D}$. The text encoder $\mathrm{TextEnc}(\cdot)$ extracts global text representation $T_{g} \in R^{D \times K}$ for each category where $K$ is the number of classes. The classification $S \in R^{K}$ is computed as follow:
\begin{align}
    &T_g = \mathrm{TextEnc}({T_K}),\\
    &V_g = \mathrm{Proj}(V_{s}),\ \ V_{s} = \mathrm{VisEnc}({I}),\\
    &S = T_g^{\mathrm{T}} V_g ,
\end{align}
where $T_g$ and $V_g$ are L2 normalized features, and $T_K$ is the sentences describing the $K$ categories. The matrix multiplication computes the cosine similarity.

\begin{figure}[t!]
  \centering
    \includegraphics[width=0.35\textwidth]{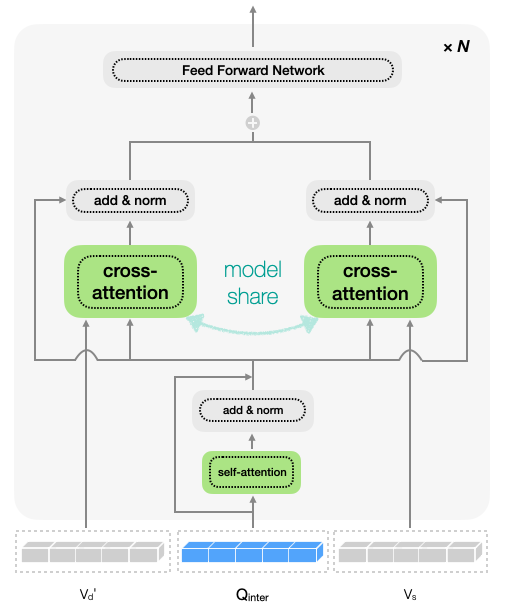}
    \vspace{-0.3cm}
    \caption{\textbf{Structure of Knowledge Integration Cross Attention.} Interaction queries first go through a self-attention layer. Then, it's fed into two shared cross-attention layers with $V_s$ and $V_d$. The outputs are summed up and fed into a feed-forward network.}
    \label{figure:fusion}
    \vspace{-0.3cm}
\end{figure}

\vspace{-4mm}
\paragraph{Interaction Decoder with Knowledge Integration} To predict HOI category for a pair of human and object queries, we generate a set of interaction queries $Q_{inter} \in R^{N_q \times C_{s}}$ by feeding the human and object features $O_h$ and $O_o$ to a projection layer. To fully exploit CLIP knowledge, we propose to retrieve interaction features from CLIP that better align with the prior knowledge in classifier weights. In detail, we preserve the CLIP spatial feature $V_{s}$ and project detection visual feature $V_{d}$ to the same dimension as $V_{s}$: 
\begin{align}
    &Q_{inter} = \mathrm{Pool}(O_o,O_h) W_i + b_i \\
    &V_d' = V_dW_p + b_p
\end{align}
where $W_i$, $b_i$, $W_p$, $b_p$ are projection parameters, $V_d'\in R^{H_s \times W_s \times C_{s}}$ and $\mathrm{Pool}$ takes average. 

To guide interaction queries $Q_{inter} \in R^{N_q \times C_{s}}$ to explore informative regions in both $V_{s}$ and $V_{d}$, we design a cross attention module for knowledge integration and its architecture is showed in Figure \ref{figure:fusion}. The $Q_{inter}$ is first updated by self-attention, and then fed into a cross-attention module with $V_{s}$ and $V_{d}'$ respectively and obtain two output features. Finally, we sum up the outputs and feed it into a feed-forward network. Formally,
\begin{align}
    &Q_{inter} = \mathrm{SelfAttn}(Q_{inter}), \\
    &C_{inter} = \mathrm{CrossAttn}(Q_{inter}, V_s), \\
    &D_{inter} = \mathrm{CrossAttn}(Q_{inter}, V_d'), \\
    &Q_{inter} = \mathrm{FFN}(C_{inter} + D_{inter})
\end{align}
where the $V_s$, $V_d'$ are the key and value respectively, and $Q_{inter}$ is the query in the shared cross attention. To extract final interaction representation $O_{inter} \in  R^{N_q \times D}$, we adopt the same projection operation as CLIP to convert the output of cross attention into the CLIP feature space as follows,
\begin{align}
&O_{inter} = \mathrm{Proj}(Q_{inter}).
\end{align}
The representation will be used for interaction classification based on a zero-shot classifier introduced in Section~\ref{sec:free}.

In this way, we leverage the object and human information from the instance decoder to retrieve interaction representation from the spatial feature map of CLIP and visual features from the detector. This query-based knowledge retrieval design allows us to achieve efficient representation learning and strong generalization capabilities.

\subsection{Verb Class Representation}
\label{sec:verb}
In this subsection, we introduce a novel pipeline to extract global verb class representation and a verb classifier built from CLIP features to cope with label imbalance.

\begin{figure}[t!]
  \centering
    \includegraphics[width=0.4\textwidth]{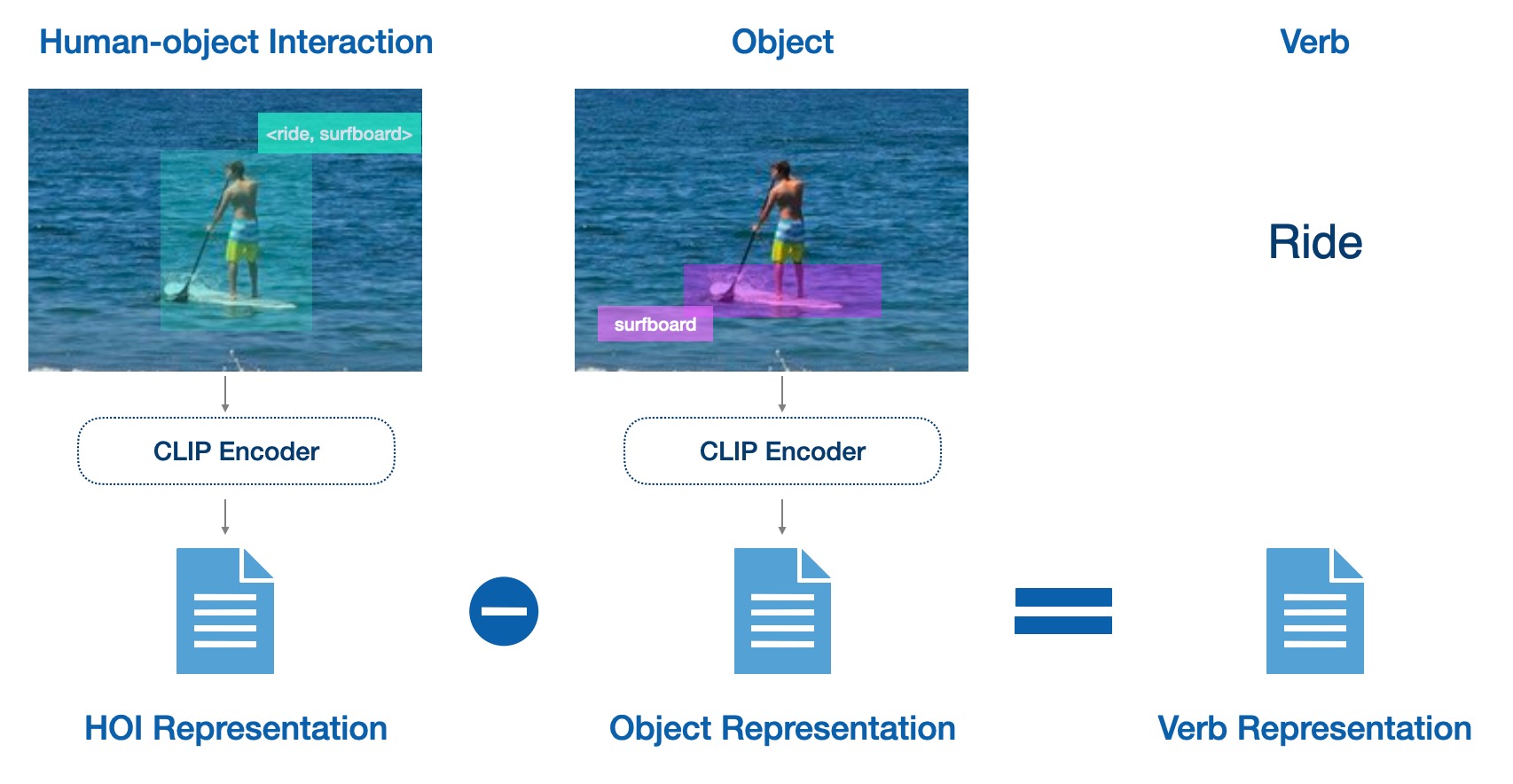}
    \caption{\textbf{Illustration of Visual Semantic Arithmetic} The object and HOI representations are extracted by encoding cropped regions of the object and HOI. Then, verb representation is obtained by HOI representation minus object representation.}
    \label{figure:arithmetic}
    \vspace{-0.6cm}
\end{figure}

\vspace{-4mm}
\paragraph{Visual Semantic Arithmetic} 
In order to better capture fine-grained verb relations from naturally imbalanced HOI annotations, we build a verb classifier through visual semantic arithmetic, which represents the global verb distribution of the training dataset. 
Here we hypothesize that the verb class representation can be derived from the difference of the global visual feature of an HOI and the global visual feature of its object. The concept is illustrated in Figure \ref{figure:arithmetic}. 

Specifically, we use the smallest region covering objects and human bounding boxes to represent an HOI triplet. Then we define $\mathbb{OBJ}_j$ as a set containing all instances of object class $j$. Additionally, we use the tuple $(i,j)$ to indicate an HOI category, where $i$ and $j$ stand for the class of verb and object respectively. Similarly, we define $\mathbb{HOI}_{(i,j)}$ as a set containing all instances of HOI category $(i, j)$.
For both HOI and object regions, we use the CLIP image encoder to obtain their visual features, then adopt a projector to map the features into a global feature space. Formally, given a region $R$, we compute its feature as follows:
\begin{align}
    f(R) = \mathrm{Proj}(\mathrm{VisEnc}(R))
\end{align}
The representation of verb class $k$ is computed by taking the difference of averaged HOI and object region features: 
\begin{align}
    &E^{k,j}_{h} = \mathrm{L2Norm}(\sum_{R_{m} \in \mathbb{HOI}_{k,j}}{f(R_{m}})) \\
    &E^{j}_{o} = \mathrm{L2Norm}(\sum_{R_{n} \in \mathbb{OBJ}_{j}}{f(R_{n}})) \\
    &E_v^k = \mathrm{L2Norm}(\sum_{n \in (k, \cdot)}{(E^{k, n}_h - E^n_o}) )
\end{align}  
where $\mathrm{L2Norm}$ stands for L2 normalization operation and $E^{k,j}_{h}$, $E^{j}_{o}$ are the computed HOI and object representations. The extracted verb class representations are prior knowledge of verb concepts from CLIP and used as verb classifier below, which are denoted as $E_v \in R^{K_v \times D}$ where $K_v$ is the number of the verb category.



\vspace{-4mm}
\paragraph{Verb Adapter} To use the verb class representation for classification, we design a light-weight adapter \cite{adapter,vladapter} module to extract verb feature $O_{verb} \in R^{N_q \times D}$ based on the interaction feature $O_{inter}$. Specifically, we use an MLP to map the interaction feature into verb feature $O_{verb} \in R^{N_q \times D}$, and compute the verb class scores as follows, 
\begin{align}
&O_{verb} = \mathrm{MLP}(O_{inter}), \\
&S_v = O_{verb} E_v^{\mathrm{T}}
\end{align}
where the verb logits $S_v$ is computed as the cosine similarity between verb feature $O_{verb}$ and verb class representation $E_v$.
In this way, we leverage the prior knowledge in the visual encoder of CLIP to extract a verb classifier from training data and design a verb adapter for better verb representation. This design generates fine-grained verb information, which benefits HOI prediction.

\subsection{Zero-shot HOI Enhancement}
\label{sec:free}
Finally, we introduce an HOI classifier generated by the prior knowledge in CLIP text encoder, which provides a training-free Enhancement for HOI classification.

Specifically, we build a zero-shot HOI classifier by exploiting the visual-linguistic alignment learned by CLIP, in which the label descriptions embedded by CLIP text encoder $\mathrm{TextEnc}$ is used as the classifier weights. Similar to\cite{eoid, gen}, we convert each HOI category to a sentence with a hand-crafted template, ``A photo of a person [Verb-ing] a [Object]". The templates are fed into the CLIP text encoder $\mathrm{TextEnc}$ to obtain an HOI classifier $E_{inter} \in R^{K_h \times D}$ where $K_h$ is the number of HOI categories. 

To leverage the zero-shot CLIP knowledge, we compute a set of additional HOI logits from the global visual feature of the image $V_{g}$ and the HOI classifier $E_{inter}$. To filter out low confidence prediction, we only keep top $K \in [0, K_h]$ scores. Formally, 
\begin{align}
&S_{zs} = \mathrm{TopK}(V_{g}E_{inter}^{\mathrm{T}})
\end{align}
where the $\mathrm{Topk}$ is the operation that select HOI logits with top $K$ score and $S_{zs}^i$ indicate score for $i^{th}$ HOI category. 
The updated $S_{zs}$ is a training-free HOI prediction with high confidence, which leverages the zero-shot CLIP knowledge to benefit tail-class prediction.

Given the zero-shot HOI classifier $E_{inter}$, we also use it to  generate an interaction prediction score based on the interaction representation $O_{inter}$ computed in Section~\ref{sec:query}, 
\begin{align}
&S_{inter} = O_{inter}E_{inter}^{\mathrm{T}},
\end{align}
which will be integrated with two other classification scores as described below.

\subsection{Inference and Training}
\label{sec:train}
In this subsection, we present the details of the training and inference pipeline of our framework.
\vspace{-3mm}
\paragraph{Training}
During training, we obtain the training HOI logits $S_{t}$ by combining HOI prediction $S_{inter}$ and verb prediction $S_v$,
\begin{align}
&S_t = S_{inter} + \alpha \cdot S_v
\end{align}
where $\alpha \in R$ is a weighting parameter. For the bipartite matching process, we follow the previous HOI detector\cite{gen,hotr,CDN,qpic} based on the DETR framework and use the Hungarian algorithm to assign ground truth to prediction. The matching cost consists of human and object bounding box regression loss, object classification loss, interaction-over-union loss, and HOI classification loss. Auxiliary losses are used on intermediate outputs of decoder layers.

\vspace{-2mm}
\paragraph{Inference}
The zero-shot HOI prediction $S_{zs}$ is used in inference time. The final HOI logits $S_{i}$ is obtained by,
\begin{align}
&S_i = S_{inter} + \alpha \cdot S_v + S_{zs}
\end{align}
Following previous methods\cite{gen}, we use the object scores $C_o$ from the instance decoder to compute the HOI triplet score, which can be written as
\begin{align}
&score^n = S_i^n + C_o^m \cdot C_o^m
\end{align}
where $n$ is the HOI category index and $m$ is the object category index corresponding with $n^{th}$ HOI category. Finally, triplet NMS is applied to top-K HOI triplets according to the confidence score.

\section{Experiments}
\label{sec:exp}
In this section, we introduce a series of experimental analyses and comprehensive ablation studies to demonstrate the effectiveness of our method. 

\begin{table}[]
\small
\centering
\begin{tabular}{llllll}
\hline
Percentage   & 100\%  & 50\%   & 25\% & 15\%  & 5\%  \\ \hline
GEN-VLKT\cite{gen}     & 33.75 & 26.55 & 22.14 & 20.40 & 15.84 \\ 
\rowcolor{blue!6}HOICLIP      &  \textbf{34.69}         & \textbf{30.88}          & \textbf{28.44}          & \textbf{27.07}          & \textbf{22.64}          \\ 
Gain(\%) &       2.96   & 16.30        & 28.46        & 32.69        & 42.92        \\
\hline
\multicolumn{6}{c}{Performance on All Categories} \\
\hline
GEN-VLKT\cite{gen}                & 29.25 & 18.94          & 14.04          & 13.84          & 13.31          \\
\rowcolor{blue!6}HOICLIP         &   \textbf{31.30}    & \textbf{26.05} & \textbf{25.47} & \textbf{24.59} & \textbf{21.94} \\
Gain(\%) &   7.00    & 37.53        & 81.41        & 77.67        & 64.84 \\ 
\hline
\multicolumn{6}{c}{Performance on Rare Categories} \\
\hline
\end{tabular}
\vspace{-0.3cm}
\caption{\textbf{Fractional Data Experiments}.}
\label{tab:frac}
\vspace{-0.6cm}
\end{table}

\subsection{Experimental setting}
\textbf{Datasets.} We conduct our experiments on two public benchmarks HICO-DET\cite{hico_det} and V-COCO\cite{vcoco}. HICO-Det contains 47,776 images (38,118 for training and 9,658 for testing). The annotations of HICO-Det conclude 600 categories of HOI triplets, which are formed from 80 object categories and 117 action categories. Among the 600 HOI categories, there are 138 categories with less than 10 training instances, defined as Rare, and the other 462 categories are defined as Non-Rare.

\textbf{Evaluation Metric.} We follow the settings of previous work\cite{hico_det, gen,qpic,CDN,XinpengLiu2022InteractivenessFI} to use the mean Average Precision (mAP) as an evaluation metric. We define an HOI triplet prediction as a true-positive example if the following criteria are met: 1) The IoU of the human bounding box and object bounding box are larger than 0.5 w.r.t, the GT bounding box; 2) the interaction category predicted is accurate.

\textbf{Zero-shot Construction.} Following previous work\cite{AnkanBansal2019DetectingHI, gen, VCL, FCL, hou2021affordance}, we construct our zero-shot setting experiments in four manners: Rare First Unseen Combination (RF-UC), Non-rare First Unseen Combination (NF-UC), Unseen Verb (UV), Unseen Object (UO) and Unseen Combination(UC). Specifically, the UC indicates that all action categories and object categories are included during training, but some HOI triplets (i.e. combinations) are missing. Under the RF-UC setting, the tail HOI categories are selected as unseen classes, while the NF-UC uses head HOI categories as unseen classes. The UV setting and UO setting indicate that some action categories and object categories are not concluded in the training set respectively. For RF-UC and NF-UC, we select 120 HOI categories as unseen classes. For UV, HOI categories involving 12 randomly selected objects among 80 object categories are defined as unseen classes. For UV, HOI categories involving 20 randomly selected verb categories are not given during training.
For UC, we follow the unseen combination setting in \cite{bansal2020detecting,shen2018scaling,ConsNet}. 
\begin{table}[]
\small
\centering
\begin{tabular}{lcccc}
\hline
Method                  & Type & Unseen         & Seen           & Full           \\
\hline
Shen et al.\cite{shen2018scaling}                  & UC          & 10.06           & 24.28          & 21.43          \\
Bansal et al.\cite{bansal2020detecting}                   & UC          & 9.18           & 24.67          & 21.57          \\
ConsNet\cite{ConsNet}               & UC          & 13.16          & 24.23          & 22.01          \\
\rowcolor{blue!6}HOICLIP                  & UC          & \textbf{23.15} & \textbf{31.65} & \textbf{29.93} \\
\hline
VCL\cite{VCL}                     & RF-UC          & 10.06           & 24.28          & 21.43          \\
ATL\cite{hou2021affordance}                     & RF-UC          & 9.18           & 24.67          & 21.57          \\
FCL\cite{FCL}                     & RF-UC          & 13.16          & 24.23          & 22.01          \\
GEN-VLKT\cite{gen}                & RF-UC          & 21.36          & 32.91          & 30.56          \\
\rowcolor{blue!6}HOICLIP$^{\dagger}$         & RF-UC         & 23.48          & 34.47          & 32.26          \\
\rowcolor{blue!6}HOICLIP                  & RF-UC          & \textbf{25.53} & \textbf{34.85} & \textbf{32.99} \\
\hline
VCL\cite{VCL}                     & NF-UC          & 16.22          & 18.52          & 18.06          \\
ATL\cite{hou2021affordance}                     & NF-UC          & 18.25          & 18.78          & 18.67          \\
FCL\cite{FCL}                     & NF-UC          & 18.66          & 19.55          & 19.37          \\
GEN-VLKT\cite{gen}                & NF-UC          & 25.05          & 23.38          & 23.71          \\
\rowcolor{blue!6}HOICLIP$^{\dagger}$         & NF-UC          & 25.71          & 27.18          & 26.88          \\
\rowcolor{blue!6}HOICLIP                  & NF-UC          & \textbf{26.39} & \textbf{28.10} & \textbf{27.75} \\ 
\hline
$\rm {ATL}^{\ast}$ \cite{hou2021affordance}              & UO             & 5.05           & 14.69          & 13.08          \\
$\rm {FCL}^{\ast}$ \cite{FCL}                & UO             & 0.00           & 13.71          & 11.43          \\
GEN-VLKT\cite{gen}                & UO             & 10.51          & 28.92          & 25.63          \\
\rowcolor{blue!6}HOICLIP$^{\dagger}$         & UO             &  9.36          & 30.32          & 26.82          \\
\rowcolor{blue!6}HOICLIP                  & UO             & \textbf{16.20} & \textbf{30.99} & \textbf{28.53} \\ 
\hline
GEN-VLKT\cite{gen}                & UV             & 20.96          & 30.23          & 28.74          \\
\rowcolor{blue!6}HOICLIP$^{\dagger}$         & UV             & 23.37          & 31.65          & 30.49             \\
\rowcolor{blue!6}HOICLIP                  & UV             & \textbf{24.30} & \textbf{32.19} & \textbf{31.09}  \\ 
\hline
\end{tabular}
\vspace{-0.3cm}
\caption{\textbf{Zero-shot performance comparison with state-of-the-art methods on HICO-DET.} 
We use RF-UC and NF-UC to represent rare first and non-rare first unseen combination settings respectively. UO is short for unseen object setting, and UV indicates unseen verb setting. $\ast$ means only the detected boxes are used without object identity information from the detector. $\dagger$ indicates HOICLIP without training-free enhancement.
} 
\label{tab:zs_results}
\vspace{-0.4cm}
\end{table}

\begin{table}[]
\tabcolsep=2pt
\small
\begin{tabular}{lcccccc}
\hline
                & \multicolumn{3}{c}{mAP Default} & \multicolumn{3}{c}{mAP Know Object}              \\ \hline
Method          & Full     & Rare    & Non-Rare   & Full           & Rare           & Non-Rare       \\ \hline
PPDM\cite{ppdm}            & 21.73    & 13.78   & 24.10      & 24.58          & 16.65          & 26.84          \\
IDN   \cite{IDN}                     & 23.36    & 22.47   & 23.63      & 26.43          & 25.01          & 26.85  \\
Zou \textit{et al.}\cite{ChengZou2021EndtoEndHO}         & 23.46    & 16.91   & 25.41      & 26.15          & 19.24          & 28.22          \\
ATL\cite{hou2021affordance}             & 28.53    & 21.64   & 30.59      & 31.18          & 24.15          & 33.29          \\
AS-Net\cite{as-net}          & 28.87    & 24.25   & 30.25      & 31.74          & 27.07          & 33.14          \\
QPIC\cite{qpic}            & 29.07    & 21.85   & 31.23      & 31.68          & 24.14          & 33.93          \\
FCL\cite{FCL}             & 29.12    & 23.67   & 30.75      & 31.31          & 25.62          & 33.02          \\
CDN\cite{CDN}             & 31.78    & 27.55   & 33.05      & 34.53          & 29.73          & 35.96          \\
SCG\cite{scg}             & 31.33    & 24.72   & 33.31      & 34.37          & 27.18          & 36.52          \\
Liu \textit{et al.}\cite{XinpengLiu2022InteractivenessFI}         & 33.51    & 30.30   & 34.46      & 36.28          & 33.16          & 37.21          \\
GEN-VLK\cite{gen}        & 33.75    & 29.25   & 35.10      & 37.80          & \textbf{34.76}          & 38.71 \\ \hline
\rowcolor{blue!6}HOICLIP$^{\dagger}$ & 34.54    & 30.71   & 35.70      & \textbf{37.96} & 34.49 & \textbf{39.02} \\
\rowcolor{blue!6}HOICLIP          & \textbf{34.69}    & \textbf{31.12}   & \textbf{35.74}      & 37.61   & 34.47  & 38.54 \\ \hline
\end{tabular}
\vspace{-0.3cm}
\caption{\textbf{Comparison with state-of-the-art methods on HICO-DET.} We use $\dagger$ to indicate the HOICLIP without training-free enhancement. All result reported are with \textbf{ResNet-50} backbone.}
\label{tab:default_hico}
\vspace{-0.6cm}
\end{table}

\textbf{Implementation Details.}
For a fair comparison with previous methods \cite{gen, qpic,CDN,XinpengLiu2022InteractivenessFI}, we use ResNet-50 as our backbone feature extractor and ViT-32/B CLIP variant. The number of layers of the transformer encoder and transformer decoder is set to 3. The number of queries is set to 64. We train HOICLIP with a batch size of 16 with optimizer AdamW and a weight decay of $10^{-4}$. The total number of training epochs is set to 90. For the first 60 epochs, we set the initial learning rate to be $10^{-4}$ and use a learning rate drop for the last 30 epochs. The parameters are initialized with MS-COCO-trained DETR. The CLIP model and classifiers are fixed during training. We conduct our experiments with a batch size of 8 on 2 NVIDIA A40 GPUs.

\subsection{Experiments on Partial Data}
To verify the data efficiency of HOICLIP, we reduce the training data to 50\%, 25\%, 15\% and 5\% respectively, and compare our performance with state-of-the-art method\cite{gen} trained with those partial data settings. We show the performance on all HOI categories and rare HOI categories respectively in Table~\ref{tab:frac} where Visual Semantic Arithmetic is performed on the whole datasets. We provide the results of partial data Visual Semantic Arithmetic in Supplementary Materials.
In each partial training data scenario, HOICLIP significantly outperforms its opponent, especially on the rare HOI categories. HOICLIP maintains a satisfactory and stable performance as the percentage of available data decreases while other methods suffer a dramatic degradation due to insufficient training data. We observe HOICLIP achieves a significant performance gain of more than 81.41\% on rare categories under $25\%$ fractional data scenario. The performance demonstrates the efficiency and robustness of our methods for knowledge transfer.


\subsection{Analysis under Regular Setting} To fully understand the characteristics of HOICLIP, we conduct experiments on various zero-shot settings, RF-UC, NF-UC, UV, and UO. We compare HOICLIP with several state-of-the-art methods. The results are shown in Table~\ref{tab:zs_results}, which demonstrates the superior performance of HOICLIP. Compared with GEN-VLKT \cite{gen}, we achieve an impressive +4.04 mAP gain under NF-UC settings for all categories and a +5.69 mAP improvement on rare categories under the UO setting. 

In addition, to further demonstrate the effectiveness of HOICLIP, we compare the performance of HOICLIP with the state of the art on the default HOI detection setting. We show the performances on dataset HICO-DET in Table~\ref{tab:default_hico}. HOICLIP achieves a competitive performance and especially a +1.87 mAP improvement in rare categories compared with GEN-VLKT. For dataset V-COCO, the results are presented in Table \ref{tab:default_vcoco}. We outperform previous methods in Scenario 1 with a new state-of-the-art performance of 63.5 AP. The improvement is less significant than that on HICO-DET given the relatively small scale of V-COCO.


\begin{table}[]
\centering
\small
\begin{tabular}{lcc}
\hline
Method    & AP(Scenario 1)  & AP(Scenario 2) \\ \hline
PMFNet\cite{PMFNet}   & 53.0   &  - \\
ConsNet\cite{ConsNet} & 53.2   &  - \\
IDN \cite{IDN}        & 53.3   &   60.3 \\
SCG\cite{scg}          & 53.4   &   60.9 \\
HOTR\cite{hotr}         & 55.2   &  64.4  \\
QPIC\cite{qpic}          & 58.8    & 61.0   \\
CDN\cite{CDN}       &  62.3  & 64.4   \\
GEN-VLKT\cite{gen}         & 62.4   &  64.4  \\
Liu \textit{et al.}\cite{XinpengLiu2022InteractivenessFI}   & 63.0   & \textbf{65.2}   \\ \hline
\rowcolor{blue!6}HOICLIP & \textbf{63.5}   & 64.8   \\ \hline
\end{tabular}
\vspace{-0.3cm}
\caption{\textbf{Performance comparison in V-COCO.}}
\label{tab:default_vcoco}
\vspace{-0.3cm}
\end{table}

\subsection{Ablation Study}

\textbf{Network Architecture Design} 
To demonstrate the characteristic of our model design, we first introduce a baseline model by following the design of GEN-VLKT but remove the knowledge distillation part, denoted as \textit{Base} in Table \ref{tab:ab_fusion}. The first modification is changing the image input feature for the interaction decoder from detection visual feature $V_d$ to visual feature $V_s$ from CLIP image encoder, denoted as \textit{+CLIP}. We observe an improvement of 2.06 mAP improvement in rare categories. Next, we replace the conventional cross-attention layer in the interaction decoder with the knowledge integration cross-attention layer and change the input to  both $V_d$ and $V_s$, denoted as \textit{+integration}. The results show \textit{+integration} outperforms all existing HOI detectors and gains +1.41 mAP compared with \textit{+CLIP}. Towards the goal of deeper interaction understanding, we add a verb adapter and verb classifier from visual semantic arithmetic, denoted as \textit{+verb}. The performance is further improved to 34.54 mAP in all categories. Finally, we introduce training-free enhancement for \textit{+verb}, denoted as \textit{+free}. A gain of +0.15 mAP is acquired without additional training. We also provide the ablation study under partial data settings in the supplementary material.


\begin{figure*}[!t]
	\centering  
    \begin{subfigure}[t]{0.18\textwidth}
         \centering
         \includegraphics[width=\textwidth]{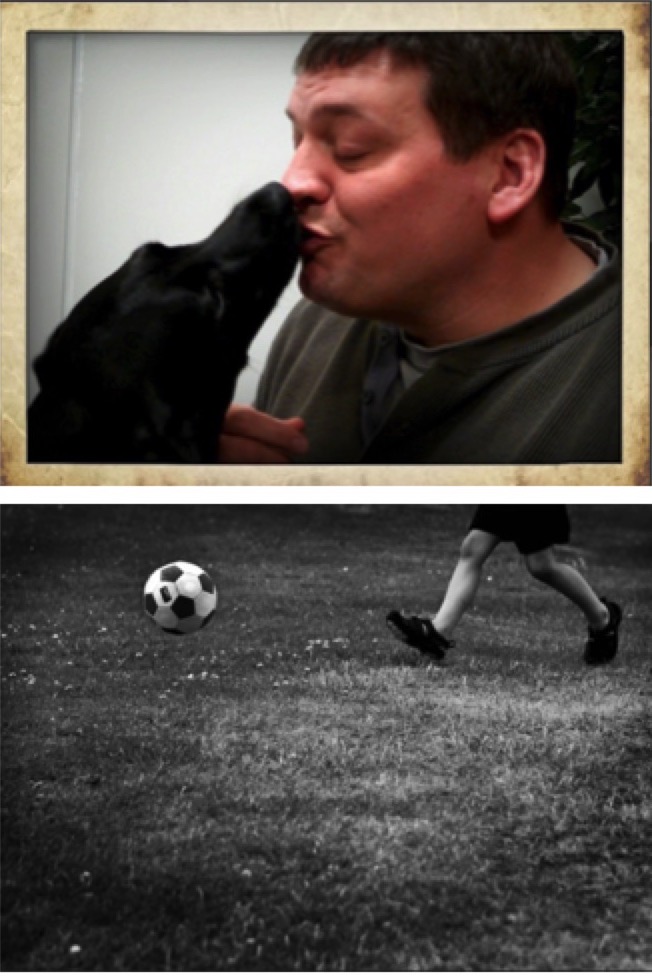}
         \caption{}
     \end{subfigure}
     \begin{subfigure}[t]{0.18\textwidth}
         \centering
         \includegraphics[width=\textwidth]{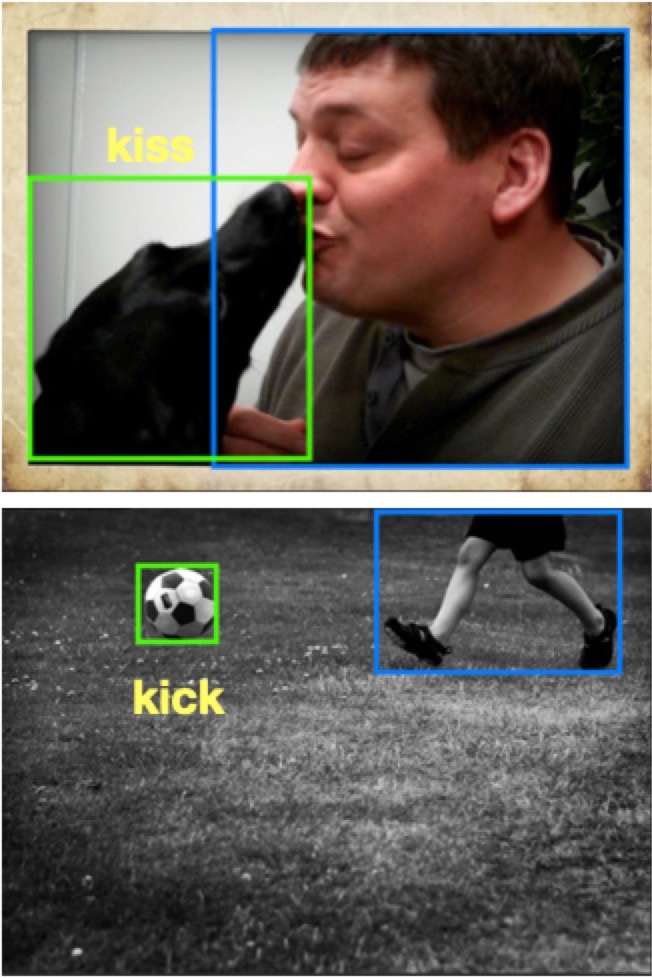}
         \caption{}
     \end{subfigure}
     \begin{subfigure}[t]{0.18\textwidth}
         \centering
         \includegraphics[width=\textwidth]{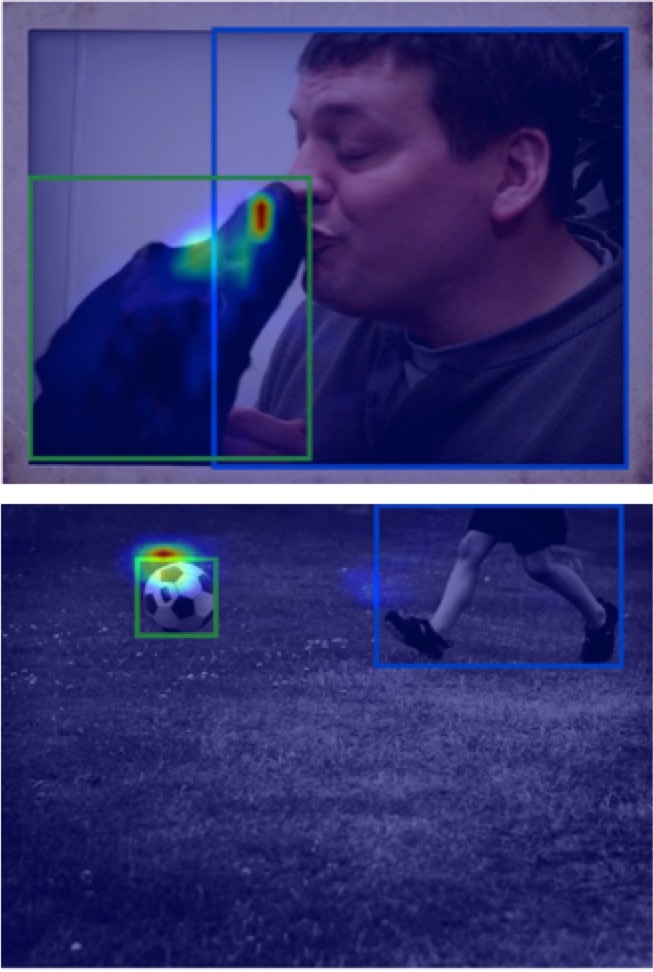}
         \caption{}
     \end{subfigure}
     \begin{subfigure}[t]{0.18\textwidth}
         \centering
         \includegraphics[width=\textwidth]{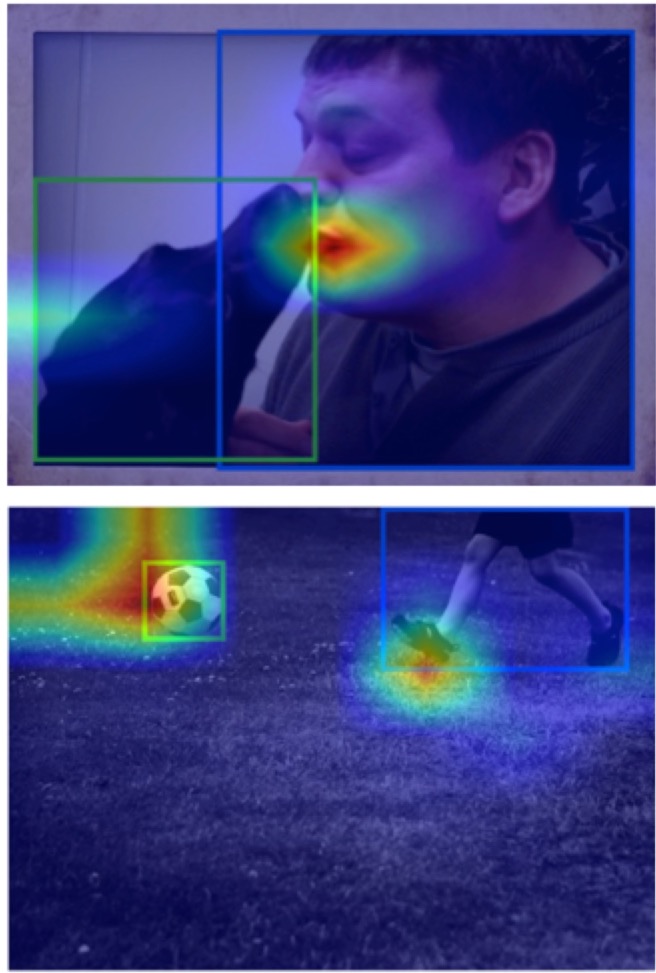}
         \caption{}
     \end{subfigure}
     \vspace{-3mm}
	\caption{\textbf{Visualization of predictions.} The columns indicate the input image $(a)$, prediction results$(b)$, attention maps from detection backbone$(c)$ and CLIP spatial feature$(d)$ in interaction decoder.}
 \vspace{-0.3cm}
    \label{fig:visualization}
\end{figure*}


\begin{table}[]
\centering
\small
\begin{tabular}{llll}
\hline
Method           & Full &Rare      & Non-rare   \\ \hline\textit{Base}            & 32.09 & 26.68 & 33.71   \\\hline
\textit{+CLIP}  & 32.72   & 28.74 & 33.92     \\ 
\textit{+integration}  & 34.13   & 30.54 & 35.20     \\ 
\textit{+verb}  & 34.54   & 30.71 & 35.70     \\ 
\rowcolor{blue!6}\textit{+free}  & 34.69   & 31.12 & 35.74     \\
\hline
\end{tabular}
\vspace{-0.3cm}
\caption{\textbf{Network Architecture Design.} Effectiveness of our components are demonstrated by accumulating each design from a \textit{Base} model.}
\label{tab:ab_fusion}
\vspace{-0.3cm}
\end{table}

\textbf{Verb Classifier Extraction}
We explore different methods to extract verb representation in the following experiments. A naive alternative is to describe human action using sentences similar to the way used in CLIP. For example, we can describe the action \textit{ride} by a sentence: ``a photo of a person riding". Having generated representation for each HOI category using CLIP, we also can obtain the representation of a verb by taking the average of all HOI representations involving the verb. For all three approaches, we use the representation of all verbs to initialize the weight of the verb adapter. We show our experiment results in Table~\ref{tab:ab_verb_weight}. We see that our visual semantic arithmetic strategy obtain a performance gain of 1.16 mAP compared with the sentence-describing method and 1.45 mAP compared with the HOI average method in the full classes setting.

\begin{table}[]
\small
\centering
\begin{tabular}{lccc}
\hline
Method                        & Full           & Rare           & Non-rare       \\ \hline
``A photo of person doing"     & 33.38          & 29.67          & 34.49          \\
Average of HOI representation & 33.09          & 28.29          & 34.52          \\ \hline
\rowcolor{blue!6}Visual semantic arithmetic    & \textbf{34.54} & \textbf{30.50} & \textbf{35.75} \\ \hline
\end{tabular}
\vspace{-0.3cm}
\caption{\textbf{Verb Classifier Extraction.} The choice of approach to generating verb representation.}
\label{tab:ab_verb_weight}
\vspace{-0.6cm}
\end{table}



\textbf{Hyperparameter Selection}
In this part, we discuss the choice of hyperparameters for each module. Different from the previous work,
we use a separate validation set, 
the details of which are provided in the supplementary material.
We train our model with different $S_v$ weights and find that the weight of $0.5$ achieves the best performance. We ablate the choice of $S_v$ weight $\alpha$ and the results are shown in Table~\ref{tab:hyper_choice}. We also choose the value of $k$ used in training-free enhancement in the same way. 
We find that $k=10$ achieves the best performance. All results are shown in Table~\ref{tab:hyper_choice}.

\begin{table}[]
\small
\tabcolsep=0.05cm
\centering
\begin{tabular}{lccclccc}
\hline
 & \multicolumn{3}{c}{Choice of $\alpha$} & & \multicolumn{3}{c}{Choice of Top $K$} \\
\hline
Weight & Full           & Rare           & Non-rare & K & Full           & Rare           & Non-rare       \\ \hline
0.00   & 31.16          & 20.96          & 34.12 & 0 & 31.65          & 22.92          & 34.18         \\
0.25   & 31.13          & 21.50          & 33.93   &5   & 32.25          & 25.14          & 34.32         \\
\rowcolor{blue!6}\textbf{0.50}   & \textbf{31.65} & \textbf{22.92} & \textbf{34.18} &\textbf{10}   & \textbf{32.26} & \textbf{24.50} & \textbf{34.51} \\
0.75   & 30.88          & 22.82          & 33.22   &15   & 32.00          & 23.79          & 34.38        \\
1.00   & 30.33          & 24.80          & 32.72 &20   & 31.96          & 23.72          & 34.35          \\ 
\hline
\end{tabular}
\vspace{-0.3cm}
\caption{\textbf{Hyperparameter Selection.} Experiments are conducted in the validation set to search for the best hyperparameters.}
\label{tab:hyper_choice}
\vspace{-0.6cm}
\end{table}




\textbf{Model Efficiency Analysis} We evaluate HOICLIP and the prior art GEN-VLKT on a single NVIDIA A100 GPU. The inference time of HOICLIP (\textbf{55.52} ms/img) is comparable to GEN-VLKT (\textbf{52.80} ms/img), and the small gap is from the additional forward through the CLIP encoder. 
Given the performance improvement, such a trade-off seems worthwhile.
The additional parameters in HOICLIP only lead to a slight increase in inference cost.
Moreover, for the model training, the number of trainable parameters in GEN-
VLKT is 129.8M under its regular setting as its CLIP module is fine-tuned while our HOICLIP has 66.1M trainable parameters due to the fixed CLIP visual encoder.

\subsection{Visualization}
We visualize the prediction result and attention maps to demonstrate the characteristics of our method in Figure \ref{fig:visualization}. The attention maps are from the cross-attention module for knowledge integration in the interaction decoder. We observe the attention map from CLIP visual feature focus on broader interaction-related regions while the attention map detection backbone only emphasizes the object regions.

\section{Conclusion}
We have presented HOICLIP, a novel framework for transferring visual linguistic model knowledge to HOI detection. HOICLIP retrieves CLIP knowledge in a query-based manner and leverages the spatial visual feature for efficient and robust interaction representation learning. We also extract a verb classifier from the prior knowledge in CLIP through visual semantic arithmetic and introduce a verb adapter for deeper interaction understanding. To further improve the model's generalization ability, we adopt a training-free enhancement for HOI classification via the text feature of CLIP. Extensive experiments demonstrate superior performance in both fully-supervised and zero-shot scenarios with high data efficiency. 


\clearpage

{\small
\bibliographystyle{ieee_fullname}
\bibliography{egbib}
}

\clearpage

\setcounter{section}{0}

\begin{center}
    \Large{\textbf{Supplementary Material}}
\end{center}


\section{Validation Set Generation}
The HICO-DET does not provide an official validation set.
When choosing our hyperparameters, we split a validation set from the training set, and also generate a new training set specially for hyperparameter selection from the remaining examples. We evaluate the performances of each hyperparameter choice on the separated validation set, to ensure fair and reliable results. We train models for hyperparameter selection on the new training set, thus model will not see the training example of the validation set during the training stage. To guarantee a proper function of the validation set and the new training set, we generate the two sets following the criterion: 1) all training instances are randomly picked from the training set; 2) there is at least one training instance for any class included in the training set. The final validation set contains 12892 images and the new training set contains 18250 images.

\section{Supplementary Analysis}

\textbf{Zero-shot HOI Enhancement for GEN-VLKT}
We conduct experiments to validate the effectiveness of zero-shot HOI enhancement in previous HOI detectors e.g. GEN-VLKT. We report the result of different top $k$ selections on test set in Table \ref{tab:gen_zs}. The training-free enhancement also works for GEN-VLKT and achieves an improvement of +0.56 mAP gain when $k$ equals to 5. We also provide the result of HOICLIP with different $k$ on the test set to explore the upper bound for zero-shot HOI enhancement in Table \ref{tab:hoi_zs}. We observe the maximum improvement is achieved when $k$ equals to 20 and we report the result with $k$ equal to 5, which is selected by the validation set.

\begin{figure}
    \centering
    \includegraphics[width=0.3\textwidth]{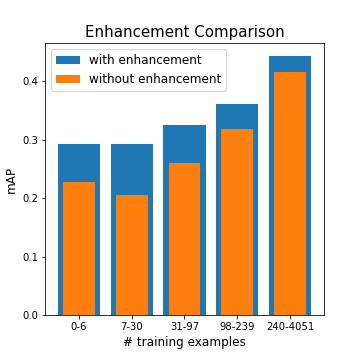}
    \caption{\textbf{Enhancement Analysis}: We split all 600 HOI categories into 5 parts by the number of training examples, then evaluate the performance of the model with/without the training-free enhancement method on each part and show the results. The performance gain is distinct especially on the classes with fewer training examples, i.e. the tail classes.}
    \label{fig:enhancement}
\end{figure}

\textbf{Visualization of Improvement}
We visualize the performance of HOICLIP with zero-shot HOI enhancement and without zero-shot HOI enhancement. The result is shown in Figure~\ref{fig:enhancement}. We observe the enhancement benefits the tail classes more compared with the head classes. We conclude the enhancement is a complement of CLIP knowledge to learned knowledge from HOI detection training data and the model with worse performance benefits more from enhancement.

\begin{table}[]
\centering
\begin{tabular}{lccc}
\hline
K &  Full     & Rare    & Non-Rare\\
\hline
0  & 33.75    & 29.25   & 35.10          \\
\textbf{5}  &\textbf{34.31}         & \textbf{30.50}          & \textbf{35.44}         \\
10 & 34.15 & 29.95 & 35.41 \\
15 & 34.16          & 30.01          & 35.40         \\
20 & 34.17          & 29.94         & 35.43          \\
25 & 34.13          & 29.88         & 35.40       \\ 
\hline
\end{tabular}
\caption{\textbf{GEN-VLKT with zero-shot HOI enhancement on HICO-DET.}}
\label{tab:gen_zs}
\end{table}

\begin{table}[]
\centering
\begin{tabular}{lccc}
\hline
K &  Full     & Rare    & Non-Rare\\
\hline
0  & 34.55    & 30.71   & 35.70          \\
5  & 34.54    & 30.71   & 35.70         \\
10 & 34.69 & 31.18 & 35.74 \\
15 & 34.71          & 31.23          & 35.74         \\
\textbf{20} & \textbf{34.75}          & \textbf{31.30}         & \textbf{35.79}         \\
25 & 34.70         & 31.11         & 35.78       \\ 
\hline
\end{tabular}
\caption{\textbf{HOICLIP with zero-shot HOI enhancement on HICO-DET.}}
\label{tab:hoi_zs}
\end{table}


\begin{figure*}[!t]
	\centering  
    \begin{subfigure}[t]{0.2\textwidth}
         \centering
         \includegraphics[width=\textwidth]{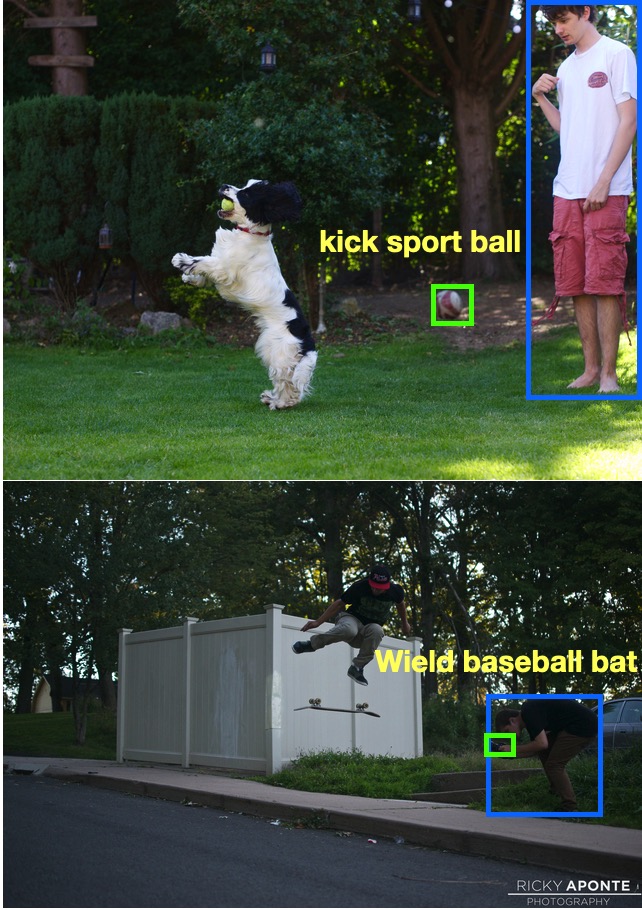}
         \caption{}
     \end{subfigure}
     \begin{subfigure}[t]{0.2\textwidth}
         \centering
         \includegraphics[width=\textwidth]{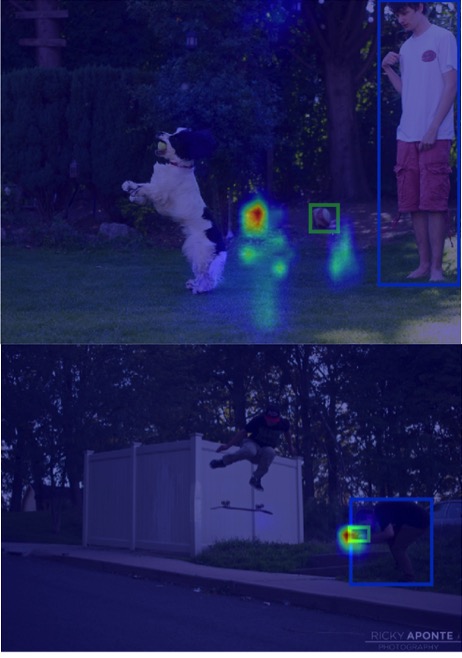}
         \caption{}
     \end{subfigure}
     \begin{subfigure}[t]{0.2\textwidth}
         \centering
         \includegraphics[width=\textwidth]{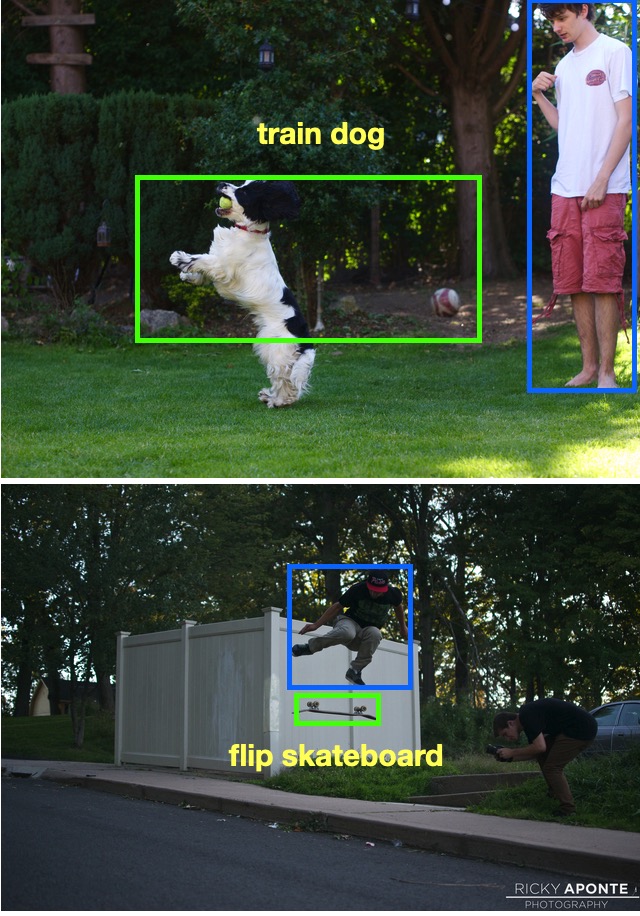}
         \caption{}
     \end{subfigure}
     \begin{subfigure}[t]{0.2\textwidth}
         \centering
         \includegraphics[width=\textwidth]{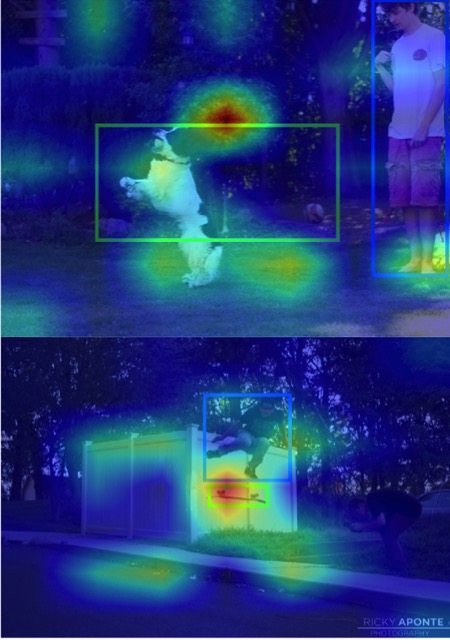}
         \caption{}
     \end{subfigure}
	\caption{\textbf{Fail cases of conventional methods.} In (a) and (b), we present the fail cases of GEN-VLKT and the corresponding attention map from the interaction decoder. In (c) and (d), we visualize the prediction from HOICLIP and the corresponding attention map from CLIP spatial feature in the interaction decoder.}
    \label{fig:visualization}
\end{figure*}

\textbf{Additional Qualitative Analysis}
We present the failed cases of conventional HOI detectors in Figure \ref{fig:visualization}. The ground truth interaction category for the first row is training a dog and flipping a skateboard for the second row. In the first row, conventional methods wrongly predict that a human is interacting with a ball instead of a dog. Meanwhile, in the second row, GEN-VLKT wrongly predicts the interaction category where a man is flipping the skateboard instead of the wield a baseball bat. As discussed in the main manuscript, we conclude the difference lies in the focus point of GEN-VLKT and HOICLIP. We observe the attention map of HOICLIP covers more informative regions and aggregates more accurate interaction information. On the other hand, GEN-VLKT simply focuses on the object region which is inconsistent with the region required for interaction prediction.

\textbf{Justification for Visual Semantic Arithmetic.}
In contrast to the previous interpretation where the verb representation is interpreted as features of the union region minus the object features, our VSA design aims to reduce the noisy cues from the object regions due to the variation in object classes. 
We verify the effectiveness of this design in Table \ref{tab:verb_result}, which shows our method \textbf{Frac} outperforms the previous representation \textbf{Union.}
2) We use a verb representation extracted from the whole dataset in the main paper experiments. To investigate the impact of using partial data, we conduct experiments in Table \ref{tab:verb_result} where \textbf{Frac} indicates verb representation extracted from only partial data settings. By comparing with {HOICLIP}, we can see that our model is robust w.r.t different amounts of data for VSA.

\textbf{Ablation Study under Low-data Regime.} We conduct an ablation study under low-data settings in Table \ref{tab:verb_result} to better demonstrate the characteristic of the proposed method. The result shows that the model relies more on the \textbf{CLIP} features with fewer data. However, all the modules work collaboratively to achieve the best performance. 

\textbf{The necessity of CLIP.} 
To verify the necessity of CLIP, we replace the CLIP visual encoder with Imagenet-pretrained ViT.
The strong performance of HOICLIP is achieved by leveraging the alignment between 
CLIP's text and visual encoders instead of using more parameters. As shown in Table \ref{tab:verb_result}, replacing CLIP ViT with Imagenet-pretrained ViT (denoted as \textbf{HOICLIP-ViT}) breaks the alignment, and the performance degrades.

\begin{table}[t]
\vspace{+0.2em}
\centering
\scalebox{0.8}{
\begin{tabular}{lcccc}
\hline
Method                  & 100\%        & 50\%           & 15\%          \\
\hline
Union             & 33.03          & 30.79           & 26.80               \\
Frac                  & 34.69          & 31.11          & 26.84        \\
\hline
\hline\textit{Base}            & 32.09 & 25.54 & 21.57   \\
\textit{+CLIP}  & 32.72   & 29.80 & 25.20     \\ 
\textit{+integration}  & 34.13   & 30.28 & 25.63     \\ 
\textit{+verb}  & 34.54   & 30.33 & 26.25     \\
\hline
\hline
HOICLIP-ViT & 33.03   & 28.28 & 23.91     \\ 
\hline
\hline
\rowcolor{blue!6}HOICLIP & \textbf{34.69}   & \textbf{30.88} & \textbf{27.07}     \\ 
\hline
\end{tabular}}
\caption{\textbf{Fractional data performance on the HICO-DET dataset.}}
\label{tab:verb_result}
\vspace{-0.2cm}
\end{table}

\section{Limitation Discussion}
We notice the numbers of training parameters in prior methods are different from HOICLIP since HOICLIP includes the CLIP visual encoder as an indivisible part of its network architecture. Specifically, during training under the default setting, GEN-VLKT  leverages CLIP as a teacher model for knowledge distillation and finetunes CLIP with a smaller learning rate while HOICLIP freezes all of CLIP parameters during training and inference. However, during inference, HOICLIP requires CLIP spatial feature which leads to additional  costs in CLIP visual encoder. In summary, HOICLIP requires less training cost but more inference cost compared with GEN-VLKT. We consider the cost in inference worthwhile for better integrating CLIP knowledge and achieving more generalized and data-efficient HOI detectors.
\clearpage


\end{document}